\documentclass{article}
\usepackage{ijcai22}

\usepackage{times}
\usepackage{soul}
\usepackage{url}
\usepackage[hidelinks]{hyperref}
\usepackage[utf8]{inputenc}
\usepackage[small]{caption}
\usepackage{graphicx}
\usepackage{amsmath}
\usepackage{amsthm}
\usepackage{booktabs}
\usepackage{algorithm}
\usepackage{algorithmic}

\usepackage{times}

\usepackage{soul}

\usepackage{amssymb}
\usepackage{booktabs}

\usepackage{enumitem}
\usepackage{float}
\usepackage{xcolor}

\usepackage[pagenumbers]{}
\title{SemiMultiPose: A Semi-supervised Multi-animal Pose Estimation Framework}

\author{
 Ari Blau$^1$\footnote{Contact Author}\and
Christoph Gebhardt$^1$\and
Andrés Bendesky$^{1}$\and
Liam Paninski$^1$\And
Anqi Wu$^2$\\
\affiliations
$^1$Columbia University\\
$^2$Georgia Tech\\
\emails
\{ari.blau, cg3161, a.bendesky, lmp2107\}@columbia.edu,
anqiwu@gatech.edu
}

\begin{document}

\maketitle
\thispagestyle{empty}
\begin{abstract}
  Multi-animal pose estimation is essential for studying animals' social behaviors in neuroscience and neuroethology. Advanced approaches have been proposed to support multi-animal estimation. 
  However, these models rarely exploit unlabeled data during training, even though real world applications have exponentially more unlabeled frames than labeled frames. 
   Manually adding dense annotations for a large number of images or videos is costly and labor-intensive, especially for multiple instances. Given these deficiencies, we propose a novel semi-supervised architecture for multi-animal pose estimation, leveraging the abundant structures pervasive in unlabeled frames in behavior videos to enhance training, particularly in the context of sparsely-labeled problems. The resulting algorithm provides superior multi-animal pose estimation results on three animal experiments compared to existing baselines, and exhibits more predictive power in sparsely-labeled data regimes.
\end{abstract}

\section{Introduction}
\label{sec:intro}

Behavior estimation from images and videos is essential in neuroscience and neuroethology. Although advanced pose estimation methods \cite{mathis2018deeplabcut,graving2019deepposekit,wu2020deep,pereira2020sleap} have achieved satisfying performance in many single-animal experiments, it is demanding to develop advanced algorithms for robust multi-animal pose estimation to study social animal behaviors. Multi-animal pose estimation is more challenging due to the following reasons:
\begin{itemize}[leftmargin=*]
    \item Compared with single-animal estimation, there is more ambiguity when there is an undefined number of animals. Body parts could have a high resemblance, both within an animal and across animals, and a low visibility. The assignment of body parts to different animals could become more difficult when animals partially overlap on top of each other and where there is no given number of animals per frame. 
\vspace{-0.05in}
    \item Compared with multi-person tracking in the human pose estimation (HPE) literature, 
    labeled dataset sizes tend to be significantly smaller in the field of animal studies. Transfer learning could well alleviate the label-deficiency issue in HPE since all humans share the same body structure, but this strategy can not be adopted across species in animal pose estimation (APE). Manually adding dense annotations for a large number of images or videos is costly and labor-intensive, especially for multiple instances. Therefore,  the label-deficiency issue remains challenging, as in many single-animal analyses \cite{wu2020deep}.
\end{itemize}

Existing multi-animal pose estimation methods can be categorized into bottom-up \cite{madlc2021,pereira2020sleap,rodriguez2018multiple} and top-down \cite{pereira2020sleap} methods. The former group of methods all rely on Part Affinity Fields (PAFs) \cite{cao2017realtime}, which identify all the possible body parts first and assembles them based on association analysis in an instance-agnostic fashion. The second-step association is non-trivial when there are many small animals with multiple similar body parts and low visibility. Such an approach involves a heuristic grouping process. Pereira et al. \cite{pereira2020sleap} proposed a top-down approach which detects instances first, and then crops the image for each instance and performs single-animal pose-estimation. These approaches cannot fully leverage the sharing computation mechanism of the backbone convolutional neural networks (CNNs). The second-step single-animal estimation would also be severely harmed if two animals are close to each other or occlude each other. Moreover, both bottom-up and top-down methods come with the price of multi-step computations. Beyond these deficiencies, most previous methods for multi-instance pose estimation in both human and animal literature rarely consider unlabeled data during training. Wu et al. \cite{wu2020deep} proposed a semi-supervised learning framework, DeepGraphPose \cite{wu2020deep}, integrating unlabeled data during training for single-animal estimation, but extending it to the multi-animal setting is highly nontrivial.

\begin{figure*}[t]
\centering
  \includegraphics[width=.95\textwidth]{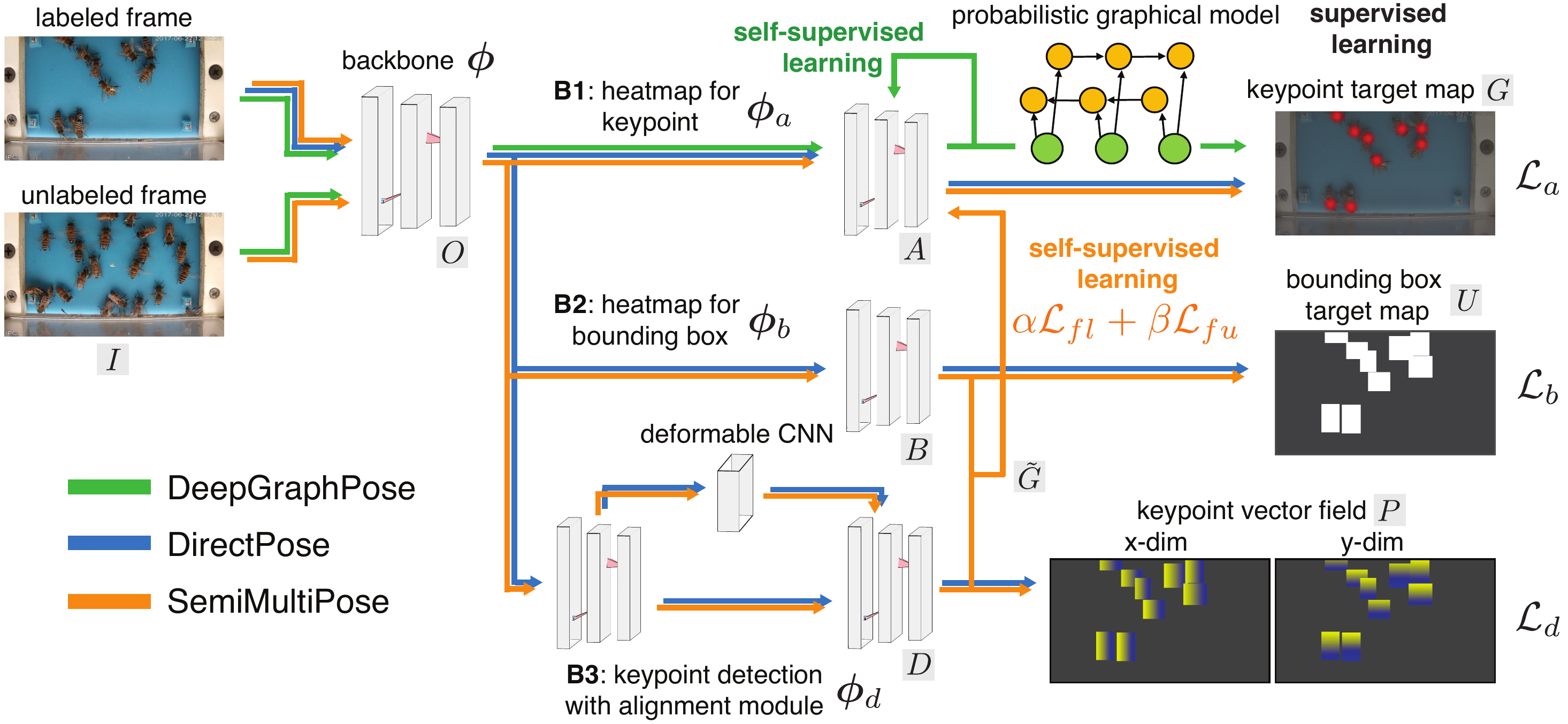}
  \caption{A comparative diagram of DeepGraphPose, DirectPose and SemiMultiPose. Methods are color-coded.}
  \label{fig:pipeline}
\end{figure*}

This paper proposes a novel semi-supervised framework, SemiMultiPose, that implements a one stage multi-animal pose estimation algorithm that is neither top-down or bottom-up and leverages the large number of unlabeled frames pervasive in animal videos (the orange pipeline in Fig.~\ref{fig:pipeline}). It is an extension of DeepGraphPose \cite{wu2020deep} to the multi-animal scenario. DeepGraphPose leverages the abundant spatiotemporal structures pervasive in unlabeled frames to enhance training, particularly in the regime of few training labels. The objective function in DeepGraphPose consists of a supervised loss for labeled frames and a self-supervised loss for unlabeled frames (the green pipeline in Fig.~\ref{fig:pipeline}). The pseudo labels in the self-supervised loss are readouts from unimodal output tensors via a soft-argmax function (continuous relaxation of hard-argmax) in the single-animal setting. The soft-argmax function guarantees the differentiablility of continuous pseudo labels for end-to-end training. However, the soft-argmax function is not applicable to multi-modal output tensors, thus hindering the readout of differentiable pseudo labels in the multi-animal setting.

We propose to resolve the issue by introducing a separate network branch that directly generates differentiable pseudo labels as 2D continuous vectors, instead of reading from multi-modal output tensors. This additional branch is adapted from DirectPose \cite{tian2019directpose}, a direct end-to-end multi-person pose estimation method (the blue pipeline in Fig.~\ref{fig:pipeline}). SemiMultiPose shares a large portion of the model architecture with DirectPose, so it inherits the benefit of DirectPose that leverages the sharing computation mechanism and sidesteps the need for heuristic grouping in bottom-up methods or boundingbox detection and cropping operations in top-down ones \cite{tian2019directpose}. The major improvements of SemiMultiPose reside in the fusion of the branches in DirectPose for self-supervised learning (Fig.~\ref{fig:pipeline}). In the experiments, we demonstrate SemiMultiPose enhances multi-animal pose estimation performance 
and exhibits more predictive power, especially in the sparsely-labeled regime.


\section{Related Work}
\paragraph{Multi-instance pose estimation.}
In HPE, both bottom-up methods \cite{cao2017realtime,newell2016associative,papandreou2018personlab,pishchulin2016deepcut} and top-down methods \cite{chen2018cascaded,fang2017rmpe,he2017mask,sun2019deep} have been extensively proposed to solve multi-person pose estimation problems. Bottom-up methods predict multiple body parts at the same time and thus are trained faster than top-down. However, the post-hoc part grouping process highly relies on effective heuristics, such as association \cite{cao2017realtime}. Top-down methods usually have better performance but they are slower due to the two-state cropping and single-instance pose estimation. There are generally two sub-types under top-down. One is anchor-based object detection \cite{liu2016ssd,redmon2018yolov3,ren2015faster} that requires pre-defined anchor boxes which are sensitive to hyperparameter selection and not flexible to detect objects with large shape variations. The other is anchor-free object detection \cite{tian2019directpose,huang2015densebox,tian2019fcos} which releases top-down methods from pre-defined anchor boxes and easily achieves end-to-end training. DirectPose \cite{tian2019directpose} is one of the state-of-the-art anchor-free object detection methods. But it doesn't include self-supervised training for unlabeled frames. In APE, several methods have been proposed to solve multi-animal tracking. Lauer et al. \cite{madlc2021} and Rodriguez et al. \cite{rodriguez2018multiple} rely on a bottom-up method, PAFs \cite{cao2017realtime}. Pereira et al. \cite{pereira2020sleap} implement both PAFs and an anchor-based two-step top-down method. 

\paragraph{Semi-supervised learning.}
Semi-supervised learning aims at improving the training of an algorithm by leveraging both unlabeled (weakly labeled) data and labeled data. It is very useful in sparsely-labeled problems. Many pose estimation algorithms have adopted such learning schemes to enhance their performances given limited training data \cite{zhang2018multiview,ukita2018semi,bertasius2019learning}. Here, we focus on semi-supervised training for multi-instance pose estimation which is underexplored. Bertasius et al. \cite{bertasius2019learning} leverage training videos with sparse annotations to learn to perform dense temporal pose propagation and estimation for multi-person pose estimation. This approach is similar to ours in terms of building a self-supervised learning mechanism for unlabeled frames. However, it heavily relies on the temporal continuity to propagate information from labeled frames to unlabeled ones. We build the self-supervised loss independent of temporal dependence so that SemiMultiPose is also applicable to images rather than just videos. In APE, DeepGraphPose \cite{wu2020deep} was proposed to leverage useful spatial and temporal constraints to propagate information from labeled frames to unlabeled frames so as to solve single-animal pose estimation in sparsely-labeled long videos. However, it cannot be trivially extended to multi-animal scenarios as stated above. We propose to combine it with DirectPose to achieve a semi-supervised learning framework for multi-animal pose estimation.
\section{Our Approach}
Conceptually, our proposed semi-supervised learning framework builds off of both DeepGraphPose and DirectPose. In this section, we first introduce the self-supervised learning idea in DeepGraphPose and demonstrate the technical bottleneck when extending it to the multi-animal scenario. Next, we introduce the DirectPose detector and explain all of the network branches for multi-instance pose estimation. Finally, we describe SemiMultiPose based on DirectPose and show how to construct the novel self-supervised learning objective to incorporate unlabeled data and thereby boost regularization, generalization, and convergence.


\subsection{DeepGraphPose}
DeepGraphPose is a semi-supervised framework enhancing pose prediction for sparsely-labeled data via leveraging unlabeled frames in behavior videos. It also includes both spatial and temporal constraints to boost information propagation among labeled and unlabeled frames. The model consists of three components: a backbone pose estimation neural network; a CNN generating 2D tensors, a.k.a. heatmaps; a probabilistic graphical model on top of heatmaps leveraging spatial and temporal information (the green pipeline in Fig.~\ref{fig:pipeline}). The objective function consists of three loss terms: (1) for labeled frames, a sigmoid cross entropy between the predicted heatmaps and the target heatmaps constructed from the {\sl ground-truth} keypoint locations; (2) for unlabeled frames,  a sigmoid cross entropy between the predicted heatmaps and the target heatmaps constructed from the {\sl pseudo} keypoint locations; (3) spatial and temporal constraints for all frames. The bottleneck resides in the second loss term. In this paper, we mainly focus on extending DeepGraphPose with the first two terms and ignoring the third term. Extending the multi-animal pose estimation with spatial and temporal constraints will be explored in future work.

Denote the input frames as $I\in \mathbb{R}^{N\times H_I\times W_I\times 3}$, where $N$ is the number of frames, $H_I$ and $W_I$ are the height and width of each frame, and $3$ indicates 3 RGB channels. The number of body parts is $K$. Denote the backbone neural network as $\phi$ and the output of the backbone as $O=\phi(I)\in\mathbb{R}^{N\times H_{O}\times W_{O}\times K}$. Then we denote the heatmap branch for keypoints as $\phi_a$ which outputs 2D heatmap tensors as $A=\phi_a(O)\in\mathbb{R}^{N\times H\times W\times K}$. For the $n$th frame and the $k$th body part, we have a 2D heatmap tensor $A_{n,k}=A[n,:,:,k]\in\mathbb{R}^{H\times W}$. Supposing the corresponding 2D keypoint coordinate is $y_{n,k}$, for both the supervised loss and self-supervsied loss, the target is to match $A_{n,k}$ to a 2D Gaussian bump centered at $y_{n,k}$ by minimizing a sigmoid cross entropy. For labeled frames, $y_{n,k}$ is a ground-truth coordinate, while $y_{n,k}$ is a pseudo coordinate read from $A_{n,k}$ for unlabeled frames. This explains why the loss for unlabeled frames is referred to as ``self-supervised" -- the coordinate it tries to regress the heatmap against comes from the heatmap itself. In DeepGraphPose, they denote the grid coordinates in $\mathbb{R}^{H\times W}$ as $\alpha_{i,j}=[i,j]$ where $i\in[1,...,H]$ and $j\in[1,...,W]$. Then the pseudo coordinate is defined via the following soft-argmax function:
\begin{eqnarray}\label{eq:softargmax}
y_{n,k} = \sum_{i,j}\alpha_{i,j}\mbox{Softmax}(A_{n,k})_{i,j}.
\end{eqnarray}
This equation ensures the differentiability of $y_{n,k}$ so that errors could be back-propagated via $A_{n,k}$ to the network parameters in $\phi_a$. However, when we have an undefined number of animals in one frame, $A_{n,k}$ becomes a multi-modal heatmap tensor and $y_{n,k}$ would be a list of 2D vectors. The soft-argmax function exploited in eq.~\ref{eq:softargmax} would not be applicable to such a multi-animal scenario. We would instead have to resort to a hard-thresholding operation over $\sigma(A_{n,k})$ to obtain a list of pseudo locations $y_{n,k}$, i.e., 
\begin{eqnarray}\label{eq:hardthreshold}
y_{n,k} = \{(i,j), \mbox{ if }\sigma(A_{n,k})_{i,j}>\delta\},
\end{eqnarray}
where $\delta$ is a pre-defined threshold value and $\sigma$ is the sigmoid function. This hard-thresholding operation blocks the differentiability of the framework. As a result, we need to resort to alternative approaches to achieve differentiable $y_{n,k}$ so as to maintain the self-supervised mechanism. 

\subsection{DirectPose}
For the multi-animal extension, we choose to adapt DirectPose to construct the pose estimation architecture. It consists of four components without the probabilistic graphical model (the blue pipeline in Fig.~\ref{fig:pipeline}): a backbone pose estimation neural network; a branch generating heatmaps to predict keypoints (B1); a branch generating heatmaps to predict bounding boxes (B2); and finally, a branch generating 2D tensors with continuous keypoint vector fields (B3). Next we elaborate these three branches. 

\paragraph{B1: heatmaps for keypoints.}

This branch is very similar to the single branch in DeepGraphPose. The network function mapping the backbone output $O$ to the heatmaps $A$ is denoted as $\phi_a$. We construct target maps for keypoints, denoted as $G\in\mathbb{R}^{N\times H\times W\times K}$. For the $n$th frame and the $k$th body part, we have a 2D target map $G_{n,k}=G[n,:,:,k]\in\mathbb{R}^{H\times W}$. $G_{n,k}$ is a multi-modal binary map. Each mode has a small square window with ones centering at the true keypoint location of a single animal; and zeros elsewhere. We can construct the loss function for B1 as 
\begin{align} \label{eq:la}
   \mathcal{L}_a=\mbox{FocalLoss}(\sigma(A),G),
\end{align}
where the sigmoid function converts the heatmap into a confidence map with values between 0 and 1. Here we choose to use focal loss instead of sigmoid cross entropy which is used in DeepGraphPose. The main advantage of using the focal loss is that it helps optimize training when there is an imbalance of positive and negative labels in the training data \cite{lin2017focal}. The focal loss is defined as
\begin{align} \label{lh}
\textrm{FocalLoss}(p, y) = \begin{cases}
-\kappa(1-p)^\gamma\log(p) & \textrm{if} \ \ y=1 \\
-(1-\kappa)p^\gamma\log(1-p) & \textrm{if} \ \ y=0 \\
\end{cases}
\end{align}
where $\gamma$ and $\kappa$ are hyperparamters.


\paragraph{B2: heatmaps for bounding boxes.}

This branch was originally introduced by \cite{tian2019fcos} to predict bounding boxes for anchor-free detectors. 
The network mapping the backbone output $O$ to the heatmaps for bounding boxes $B\in\mathbb{R}^{N\times H\times W\times K}$ is denoted as $\phi_b$. We construct target maps for bounding boxes, denoted as $U\in\mathbb{R}^{N\times H\times W\times K}$. For the $n$th frame and the $k$th body part, we have a 2D target map $U_{n,k}=U[n,:,:,k]\in\mathbb{R}^{H\times W}$, a binary map with ones within the bounding boxes and zeros outside. We define the minimum enclosing rectangles of keypoints of the instances as pseudo bounding boxes. The loss term for B2 is
\begin{align} \label{eq:lb}
   \mathcal{L}_b=\mbox{FocalLoss}(B,U).
\end{align}
We will use B2 to generate differentiable keypoint coordinates together with B3. 

\paragraph{B3: keypoint detection.}

This is a new branch constructed in DirectPose. The branch network is denoted as $\phi_d$ that maps $O$ to $D\in\mathbb{R}^{N\times H\times W\times K \times 2}$ where $2$ corresponds to the x-dimension and y-dimension. We construct the target tensor for keypoint detection $P\in\mathbb{R}^{N\times H\times W\times K \times 2}$. For the $n$th frame, the $k$th body part and the x-dimension, we have a 2D target tensor $P_{n,k,x}=P[n,:,:,k,1]\in\mathbb{R}^{H\times W}$. Each element in $P_{n,k,x}$ represents the vector pointing from the element to the closest keypoint along the x-dimension; the same along the y-dimension. The loss term is written as
\begin{align} \label{eq:ld}
   \mathcal{L}_d=\mbox{MSE}(D\odot U, P\odot U),
\end{align}
where $U$ is the binary map for pseudo bounding boxes and $\odot$ denotes element-wise matrix multiplication. This means that we are cropping two boxes of the same size around the ground truth keypoints ($P$) and the predicted keypoints ($D$), and we are comparing those two cropped boxes. Given the loss expression, we can tell that the loss: 1) only focuses on the coordinates within the bounding boxes; 2) encourages $D$ to encode the 2D vector field of keypoints provided by $P$. By saying ``vector field", we are picturing a 2D vector space with each dot pointing to its closest keypoint. 

\paragraph{Supervised learning objective function.}

The final objective function in DirectPose is written as 
\begin{align} \label{eq:sl}
   \mathcal{L}_{supervised}=\mathcal{L}_a+\mathcal{L}_b+\mathcal{L}_d.
\end{align}
All the branches share the same size of output tensors, i.e., $H\times W$. This provides shape consistency to enforce the fusion of these branches both in eq.~\ref{eq:ld} and the following equations in SemiMultiPose. SemiMultiPose shares the same structure as in DirectPose. We thus skip the elaboration on the parameterization of $\{\phi, \phi_a,\phi_b,\phi_d\}$ which can be referred to \cite{tian2019directpose}. $\{A,B,D\}$ are network outputs and $\{G,U,P\}$ are known structures constructed from ground-truth keypoints. We can combine B2 and B3 to readout differentiable pseudo keypoint locations which unblocks the bottleneck in eq.~\ref{eq:hardthreshold}.

\begin{table*}[t]
  \centering
  \resizebox{16cm}{!}{
  \begin{tabular}{l|lll|l|l}
    \toprule
    Dataset  & \multicolumn{3}{c|}{Honeybee ($\sigma^2=1$)}  & Mouse pup ($\sigma^2=1$) & Fly ($\sigma^2=0.025$) \\
    \midrule
    Method     & AP-135  & AP-25  & AP-5 & AP & AP\\
    \midrule
    Sleap bottom up & $0.811$  & $0.779$  & $0.662$ & $0.692$ & $0.810$ \\
    Sleap top down &$0.793$ &  $0.724$ & $0.690$ & $0.620$ & $0.734$ \\
    Directpose & $0.846\pm 3.19e^{-3}$ & $0.819\pm 4.26e^{-3}$ & $0.734 \pm 3.10e^{-2}$ & $0.715 \pm 4.17e^{-3}$ & $0.794 \pm 6.52e^{-3}$\\
    SemiMultiPose-labeled   & $0.847 \pm 1.67e^{-3}$ &  $0.824 \pm 3.70e^{-3}$ &  $0.743  \pm  2.59e^{-2}$&  $0.733 \pm 4.87e^{-3}$ & $0.801 \pm 6.29e^{-3}$  \\
    SemiMultiPose    & ${\bf 0.853\pm9.62e^{-4} }$ & ${\bf 0.837 \pm 3.89e^{-3}}$  &  ${\bf 0.776 \pm 1.45e^{-2}}$ & $\bf{0.754 \pm 5.48e^{-3}}$& ${\bf 0.815\pm 2.90e^{-3}}$ \\
    \bottomrule
  \end{tabular}}
  \caption{Test AP for the three datasets. The first row indicates dataset. For honeybee, the second row indicates AP values for the different training sets. AP values must fall in the range of 0 to 1. A larger AP means that the performance is better. We see that SemiMultiPose (bold) has the highest AP for all of the datasets, which means it is the best model tested. We label the $\sigma^2$ value for each dataset. The $\sigma^2$ value for fly is smaller since the bodyparts of the flies are tiny compared with the size of the image. We need to have a small tolerance for the annotator uncertainty used in defining OKS. }
  \label{bee_map}
  \vspace{-0.15in}
\end{table*}

\subsection{SemiMultiPose}
As we can see, DirectPose is only applicable to labeled data where all frames have ground-truth keypoint coordinates. We propose SemiMultiPose by extending the loss function (eq.~\ref{eq:sl}) with additional terms that inhert the same self-supervised learning idea as in DeepGraphPose. 

\paragraph{Fusion of B1 and B3.} 
The idea is to generate pseudo keypoint coordinates from B2 and B3, and use them for the self-supervised branch looping back to B1 (Fig.~\ref{fig:pipeline}). We are essentially computing the heatmap loss between our predicted keypoints generated from B2 and B3 and the heatmap generated for that image (via B1). The predicted heatmap for bounding boxes from B2 is $B$. For the $n$th frame and the $k$th body part, the predicted 2D tensor is $B_{n,k}\in\mathbb{R}^{H\times W}$. The sigmoid function of $B_{n,k}$, i.e., $\sigma(B_{n,k})\in[0,1]$, represents the confidence level of each pixel belonging to a box. We then perform the hard-thresholding of $\sigma(B_{n,k})$ with a pre-defined threshold $\delta$ to collect a list of 2D indices $s$ and their corresponding confidence values $v\in[0,1]$. The list of 2D indices corresponds to the pixels that are most likely within some bounding boxes. Now we turn to B3 and its output tensor $D$. Again, let $D_{n,k}\in\mathbb{R}^{H\times W\times 2}$ represent the tensor for the $n$th frame and the $k$th body part. Remember that $D_{n,k}$ should encode the 2D vectors pointing from each element on the grid of the tensor to its closest keypoint. Therefore, $\tilde{D}_{n,k}=D_{n,k}+M$, where $M\in\mathbb{R}^{H\times W\times 2}$ represents the 2D coordinates on the grid for the x and y dimensions, gives us the absolute locations of the predicted keypoints. Finally, the pseudo locations would be a list of selected values from $\tilde{D}_{n,k}$ indexed by $s$, denoted as
\begin{align} \label{eq:soft_y}
   y_{n,k}=\tilde{D}_{n,k}(s),
\end{align}
for the $n$th frame and the $k$th body part, and their corresponding confidence values are $v$. We will unavoidably have more pseudo keypoints than the true number of keypoints, but many should have low confidence values. We resolve the issue by providing 
a score-value-weighted focal loss defined as $\textrm{FocalLoss}_v(p,v) = -v\kappa(1-p)^\gamma\log(p)-(1-v)(1-\kappa)p^\gamma\log(1-p)$. 
The fusion loss can be written as
\begin{align} \label{eq:la_fusion}
   \mathcal{L}_f=\mbox{FocalLoss}_v(\sigma(A),\tilde{G}),
\end{align}
where $\tilde{G}$ is the target map built from the pseduo keypoints, which is a function of $y_{n,k}$ and $v$. This loss ensures that we don't under-penalize the pseudo points with low confidence nor over-estimate the number of instances. Note that we derive $s$ and $v$ from a hard-thresholding operation on $B$ which leads to non-differentiability through $B$, whereas the values of the pseudo keypoint locations $y_{n,k}$ originate from $D$ via differentiable operations. Eq.~\ref{eq:soft_y} is the resolution for eq.~\ref{eq:hardthreshold}.

\paragraph{The fusion loss for labeled frames.} 
We first formulate the fusion loss for labeled frames only, denoted as $\mathcal{L}_{fl}$. This appears to be a meaningless term given that we already match the heatmap tensor $A$ to the target map $G$ via minimizing $\mathcal{L}_a$ (eq.~\ref{eq:la}). However, we observe that in practice the predictions from B1 and B3 are highly inconsistent in the original DirectPose model, in spite of the alignment of both branches to the same true keypoints. To enhance the consistency and information flow in-between B1 and B3, we impose the fusion loss $\mathcal{L}_{fl}$ for labeled frames.

\paragraph{The fusion loss for unlabeled frames.} 
Ultimately, we are able to construct the fusion loss for unlabeled frames, denoted as $\mathcal{L}_{fu}$.
This is the only term involving unlabeled frames. It encourages learning from structured information in massive unlabeled frames and significantly improves the predictive power over sparsely-labeled datasets. 

\paragraph{Semi-supervised learning objective function.}
The final objective function for SemiMultiPose is written as 
\begin{align} \label{eq:ssl}
 \!\!\!\!\!  \mathcal{L}_{semi-supervised}&=\mathcal{L}_{supervised}+\alpha\mathcal{L}_{fl}+\beta\mathcal{L}_{fu}\nonumber\\
&=\mathcal{L}_a+\mathcal{L}_b+\mathcal{L}_d+\alpha\mathcal{L}_{fl}+\beta\mathcal{L}_{fu},
\end{align}
where $\alpha$ and $\beta$ are two tunable hyperparameters balancing the trade-off among these loss terms. The last two terms both construct losses from pseudo keypoints, thus referred to as the self-supervised learning terms. The inference of SemiMultiPose is via $y_{n,k}$ and $v$. We select the pseudo keypoints in $y_{n,k}$ and weigh them by confidence values $v$.

\section{Discussion.}
Here we provide some core motivations of the branches in the proposed framework and thus help to understand our SemiMultiPose model:
\vspace{-0.05in}
\begin{itemize}[leftmargin=*]
    \item In the original DirectPose paper, the authors predict keypoints during test time via B3 only. But they still propose to train the model with the heatmap loss ($\mathcal{L}_a$) for the reason that: the joint learning of a keypoint-based regression task (eq.~\ref{eq:ld}) and a heatmap-based task (eq.~\ref{eq:la}) can be used to further boost the precision of keypoint localization. In our proposed framework, we will show that a fusion of these two branches (B1 and B3) can not only enhance the predictive power but also improve the convergence performance. Therefore, the heatmap-based branch (B1) is essential to train a better model.
\vspace{-0.05in}
    \item The heatmap-based branch for bounding box prediction (B2) is indispensable for two reasons: 1) 
    B2 doesn't help much with accurate keypoint localization but instead boosts the precision of object localization with bigger objects to detect. This provides a robust object/anchor detection that regularizes the training of the backbone network, and
    2) B3 doesn't incorporate the bounding box information during training.
    But the fusion loss as well as the inference procedure depend on both B2 and B3. 
\vspace{-0.05in}
    \item The branch function $\phi_d$ in B3 has an intricate keypoint alignment module proposed by DirectPose composed of two steps. The first step finds a rough estimation of keypoint locations for each instance; the second step employs a deformable convolutional network \cite{dai2017deformable} that allows local warping of the first-step coarse locations to well align the target keypoints. This explains why DirectPose-based methods could outperform many bottom-up and top-down methods: they predict keypoints with strong spatial association (bounding box and the first-step keypoint detection); and they increase the network capacity with the additional deformable CNN that functions as a local refinement module improving keypoint localization drastically. Further details can be found in \cite{tian2019directpose}.
\end{itemize}

\section{Experimental Results}\label{sec:exp}
We conduct experiments on three animal datasets: honeybees\footnote{Example data for honeybees can be found at \url{https://github.com/piperod/beepose/}.  BSD-3-Clause License.} \cite{rodriguez2018multiple}, mouse pups\footnote{Data unpublished.} and flies\footnote{Example data for flies can be found at \url{https://github.com/talmo/sleap-mit-tutorial}. Creative Commons Zero v1.0 Universal license.} \cite{pereira2020sleap}. We compare our SemiMultiPose model against two supervised learning approaches, DirectPose and SLEAP (both bottom-up and top-down implementations). DirectPose is the multi-person pose estimation baseline model we develop our model upon. SLEAP implements a state-of-the-art multi-animal pose estimation algorithm.

\paragraph{Implementation details.}
We use ResNet-50 \cite{he2016deep} as the backbone network. The weights are initialized from pre-trained ImageNet weights \cite{deng2009imagenet}. The number of iterations is 15k for DirectPose and SemiMultiPose for the honeybee and fly data, and 40k for the mouse pup data.
The batch size for each iteration is 3. The models are trained with stochastic gradient descent (SGD). We set the initial learning rate to 0.01 and decay it by a factor of 100 after 5k iterations. 
Other implementation details can be found in \cite{tian2019directpose}. The hyperparameters are pre-defined as follows: $\gamma=2$ and $\kappa=0.25$ for the focal loss; $\delta=0.05$ for the hard-thresholding on the bounding box heatmaps; $\alpha=0.01$ and $\beta=0.1$ for the semi-supervised loss function in eq.~\ref{eq:ssl}. We choose these values because they consistently render the best empirical performance over all datasets. 
In cases where there is a significant lack of labeled data available, we increase $\alpha$ to the range of $5$ to $10$. The intuition is that when there is less labeled data, we want the fusion loss for the labeled frames ($\alpha \mathcal{L}_{fl}$) to be stronger since it helps with regularization. SLEAP uses 200 epochs with an early stopping mechanism and is trained with the Adam optimizer. An epoch is a full iteration over the training dataset. The batch size is 4. The initial learning rate is $1 \times 10^{-3}$ which is reduced by a factor of 0.5 after 5 epochs without a minimum validation loss decrease of $1 \times 10^{-6}$, followed by a 3 epoch cooldown period.

\begin{figure*}[h]
\centering
  \includegraphics[width=.85\linewidth]{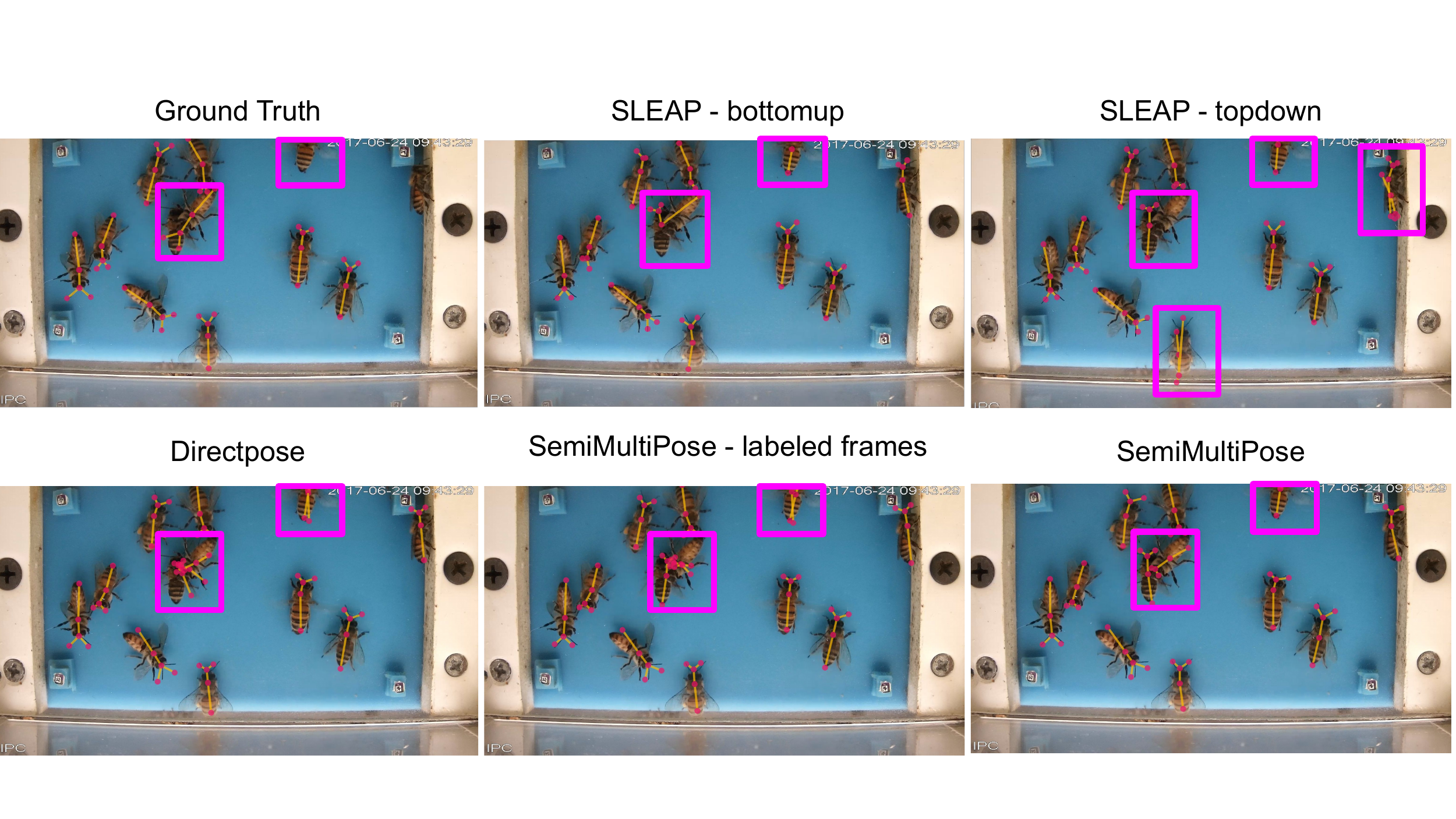}
  \vspace{-0.15in}
  \caption{Qualitative comparison of SLEAP-bottomup, SLEAP-topdown, DirectPose, SemiMultiPose-labeled and SemiMultiPose on a test frame from the honeybee data. The ground truth is presented in the first figure. The model is trained with Train5. The purple boxes highlight where the methods have different predictions and/or where there are erroneous predictions. Note that the ground truth markers themselves are not completely accurate. By comparing the bees in the boxes, we see that SemiMultiPose has the most accurate predictions.}
   \vspace{-0.05in}
 \label{fig:compare_bee}
\end{figure*}

\paragraph{Honeybee dataset.} 
The video capture system is designed to observe the ramp through which all foraging bees must pass to exit or enter the colony. Each bee has five body parts on its skeleton: tail, thorax, head, left antenna, and right antenna. The dataset consists of 165 fully annotated frames and 80 unlabeled frames, where each frame contains from 5 to 25 individual bees. We split 165 labeled frames into a 135-frame training set (Train135) and a 30-frame test set. We also create ten additional training sets, five with 25 frames (Train25) and five with 5 frames (Train5) respectively, randomly selected out of the same 135-frame training set. The two smaller training set sizes will be used to demonstrate the necessity of incorporating unlabeled frames for sparsely-labeled problems. The 80-frame unlabeled set for semi-supervised training is consistent across different training sets. Additionally, for Train25 and Train5, for each dataset, we use the images not in the data sets as supplemental unlabeled images without their annotations. Fig.~\ref{fig:compare_bee} shows a comparison of SLEAP-bottomup, SLEAP-topdown, DirectPose, SemiMultiPose-labeled and SemiMultiPose for Train5. SemiMultiPose-labeled is SemiMultiPose fed with labeled frames only. This is run to reveal the benefit brought by $\mathcal{L}_{fl}$. Comparing to both DirectPose and SemiMultiPose-labeled, we can understand the full objective function (eq.~\ref{eq:ssl}) for SemiMultiPose more comprehensively. We mark the differences between methods with purple rectangles in Fig.~\ref{fig:compare_bee}. We see that none of the methods can correctly identify the two bees that are overlapping in the center besides SemiMultiPose. SemiMultiPose has overall the best pose prediction on all bees in this frame. We also include video comparisons of these methods for all test frames at \href{https://drive.google.com/drive/folders/1hgggN9swuahsfr2Lsh6yRkRmzpx8a-VG?usp=sharing}{\textcolor{blue}{this link}} ({https://drive.google.com/drive/folders/1hgggN9swuahsfr2Ls
h6yRkRmzpx8a-VG?usp=sharing}).

\begin{figure*}[h!]
\centering
   \includegraphics[width=1\linewidth]{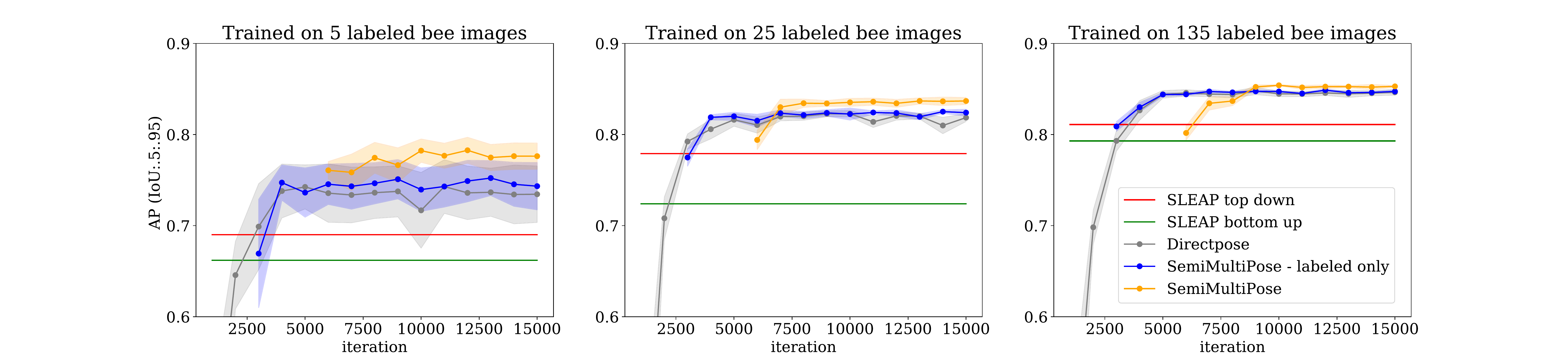}
  \vspace{-0.05in}
  \caption{Test AP for the honeybee dataset. The methods for comparison are color-coded. We show the comparison of test AP curves on three different sizes of training set. The shadow area indicates standard errors. In all of these graphs, SemiMultiPose (yellow) has the highest test AP, which means SemiMultiPose has the best performance.}
  \label{fig:curve_bee}
\end{figure*}

\begin{figure*}[h!]
\centering
  \includegraphics[width=.8\linewidth]{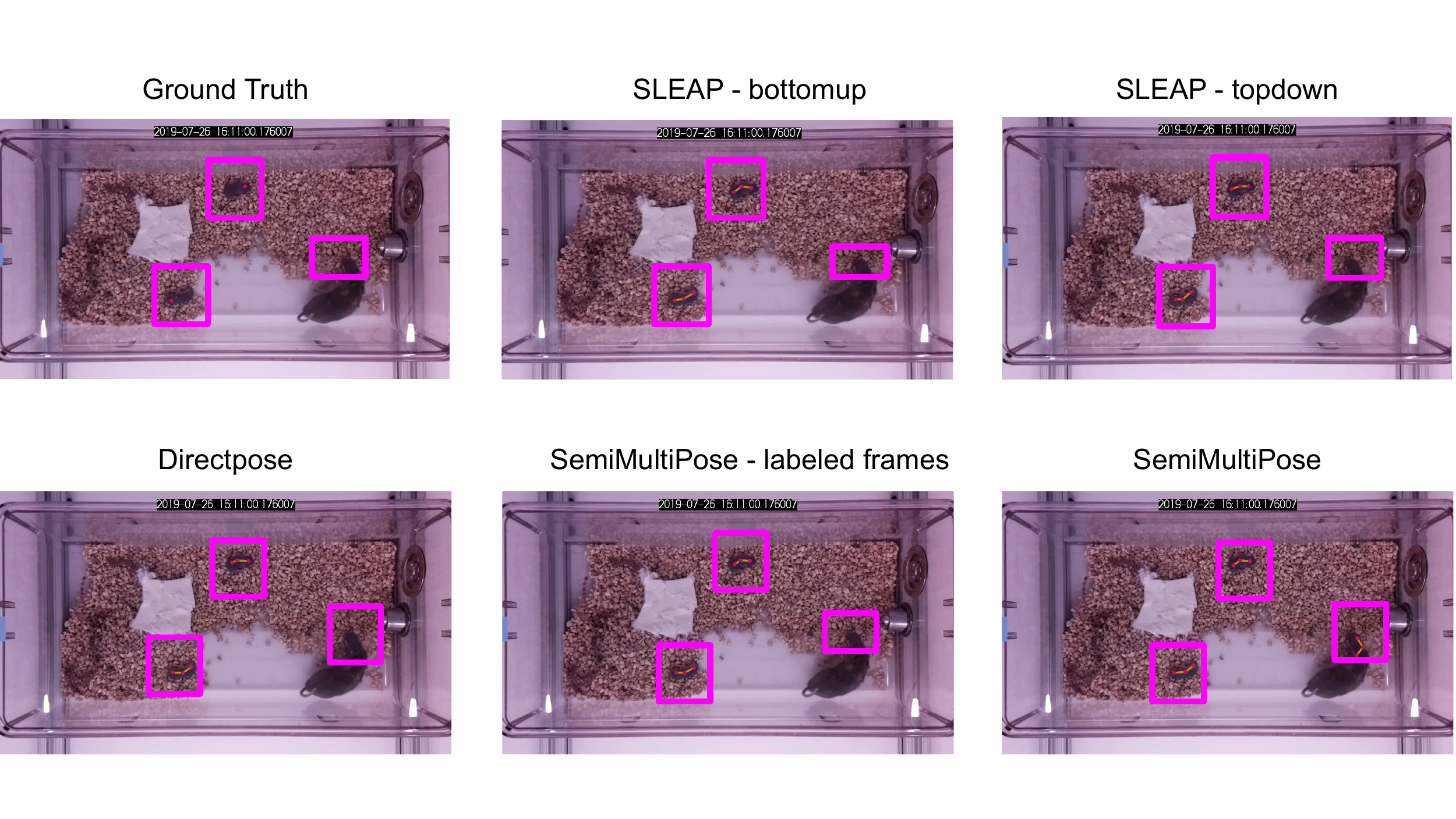}
  \caption{Qualitative comparison of SLEAP-bottomup, SLEAP-topdown, DirectPose, SemiMultiPose-labeled and SemiMultiPose on a test frame from the pup data. Conventions as in Fig.~\ref{fig:compare_bee}. We see that SemiMultiPose is the only method to correctly identify all three pups in the image.}
  \label{fig:compare_pup}
\end{figure*}

\begin{figure*}[h!]
\centering
  \includegraphics[width=.7\linewidth]{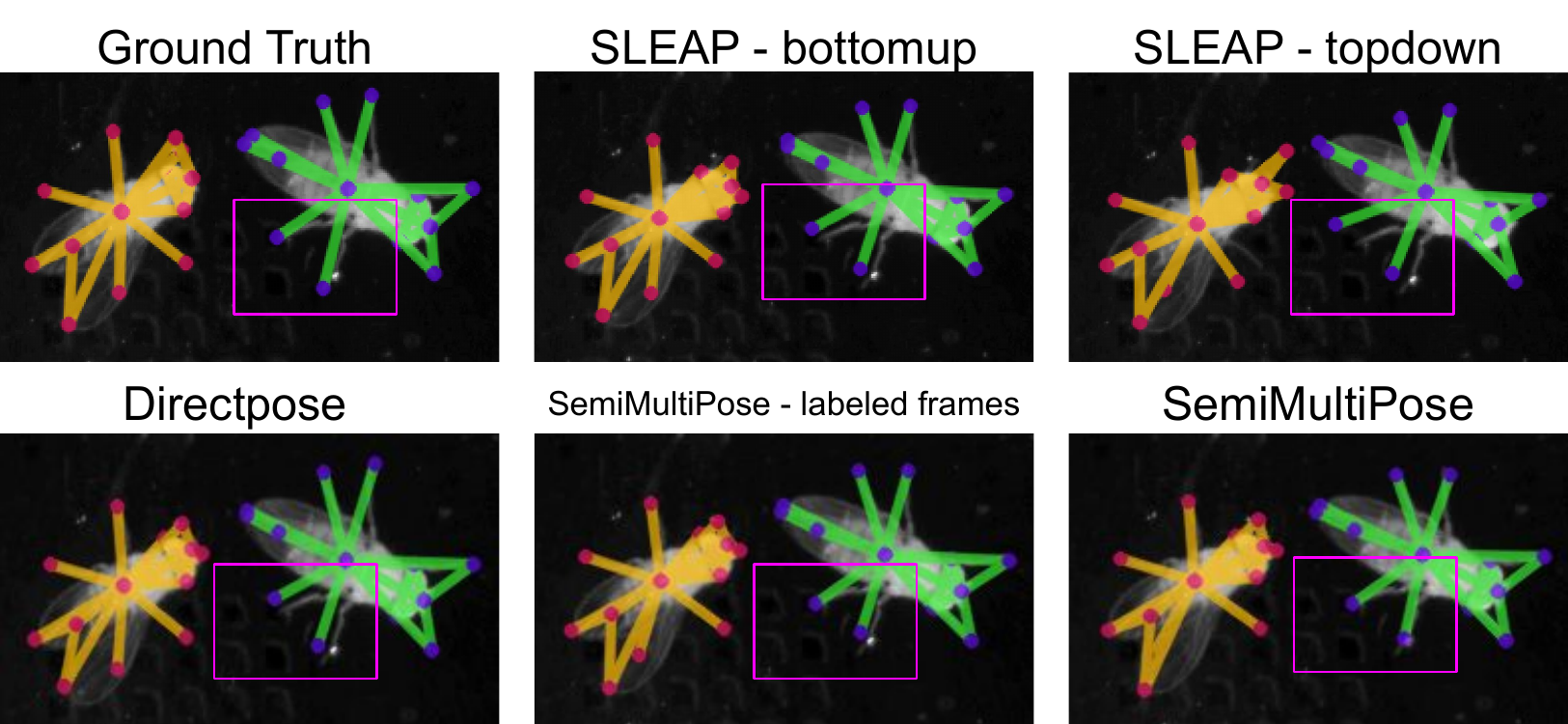}
  \caption{Qualitative comparison of SLEAP-bottomup, SLEAP-topdown, DirectPose, SemiMultiPose-labeled and SemiMultiPose on a test frame from the fly data. Conventions as in Fig.~\ref{fig:compare_bee}. We see that SemiMultiPose has the most accurate predictions for the legs of the fly on the right.}
  \label{fig:compare_fly}
\end{figure*}

\begin{figure*}[h!]
\centering
\begin{minipage}{.5\textwidth}
  \centering
  \includegraphics[width=1\linewidth]{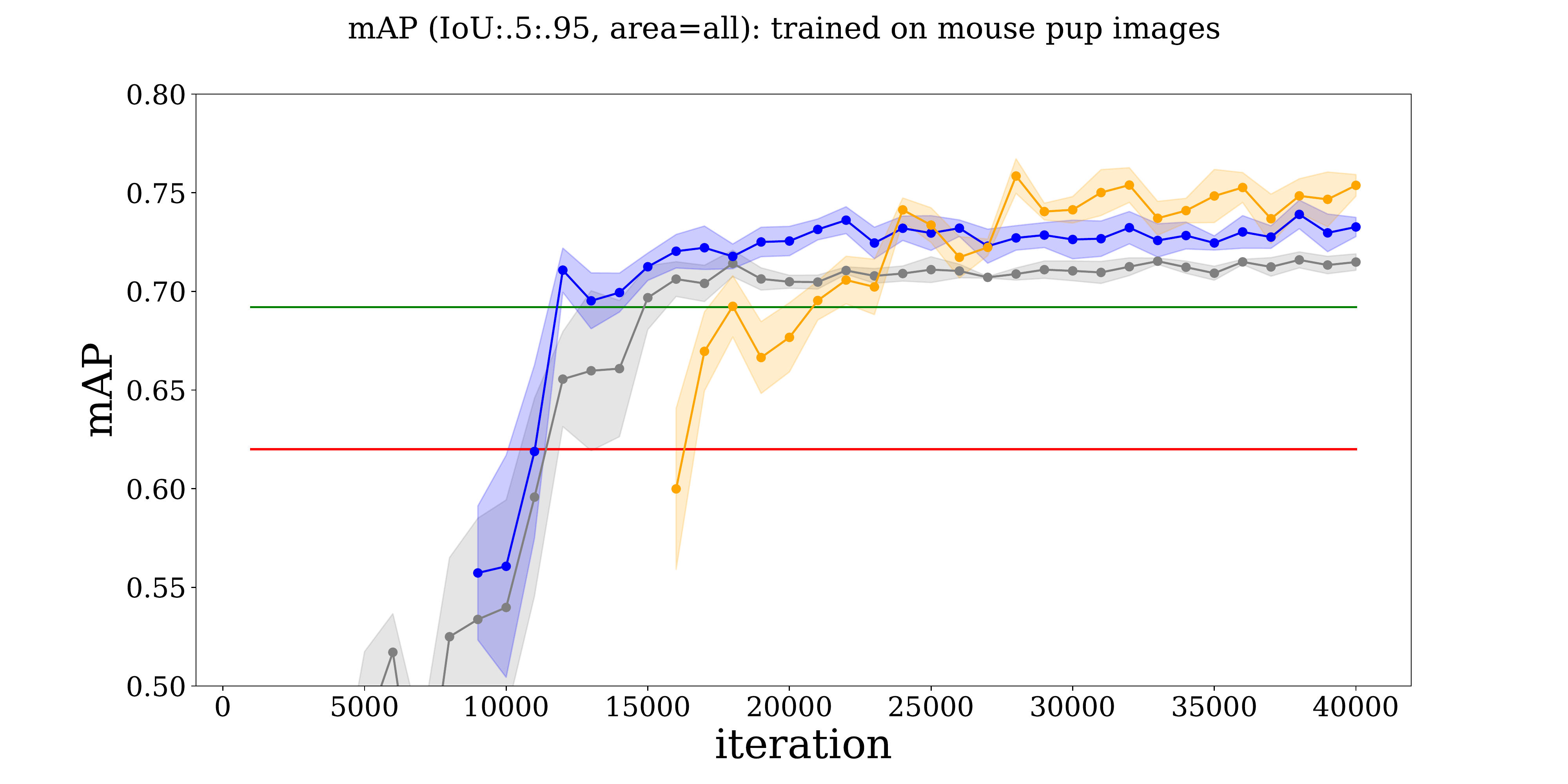}
  \caption{Test AP for the mouse pup dataset. }
  \label{fig:curve_pup}
\end{minipage}%
\begin{minipage}{.5\textwidth}
  \centering
  \includegraphics[width=1\linewidth]{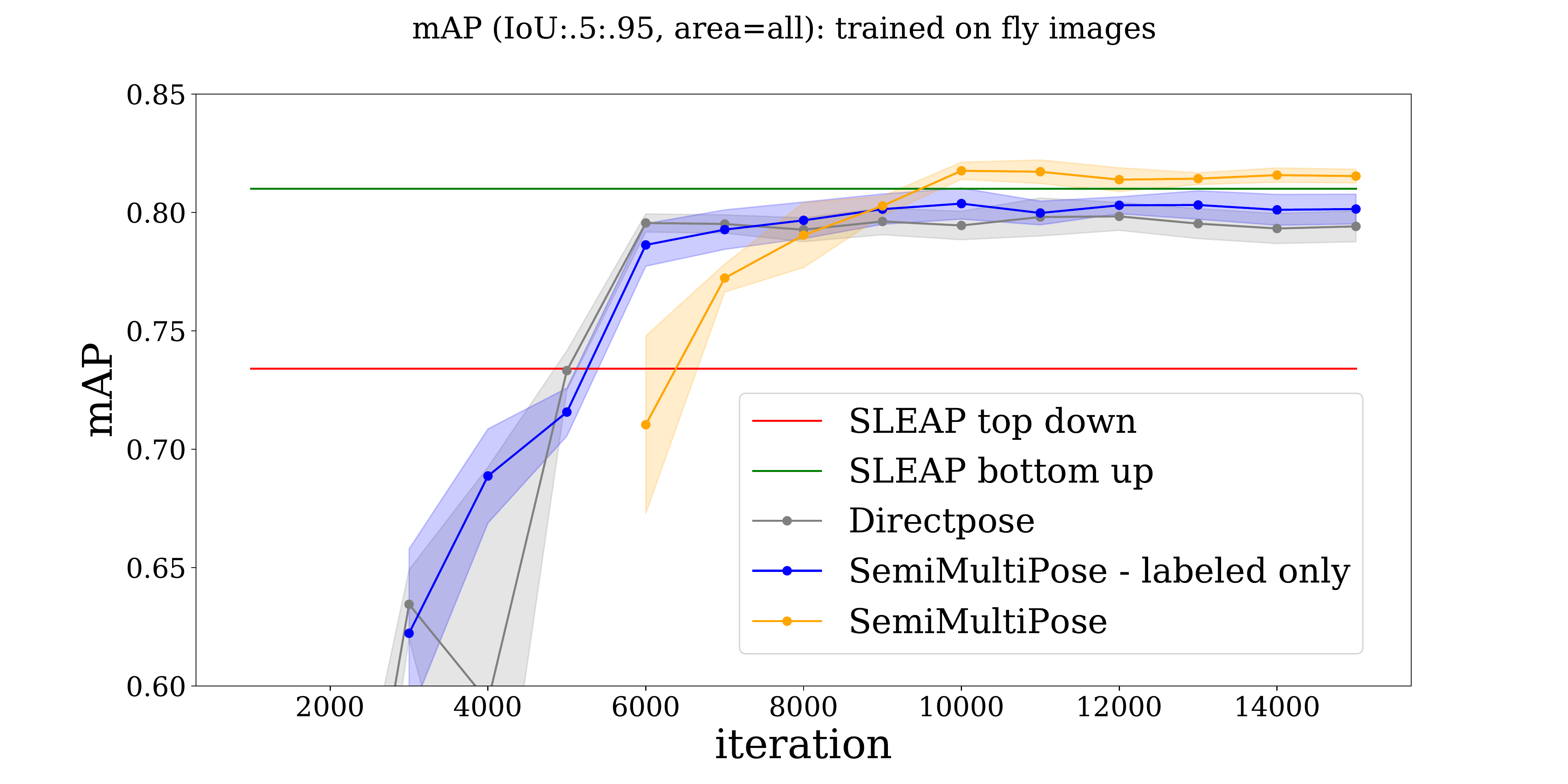}
  \caption{Test AP for the mouse fly dataset. }
  \label{fig:curve_fly}
\end{minipage}
\end{figure*}

  


Next, we present the comparison among the same methods quantitatively (Table~\ref{bee_map}, Fig.~\ref{fig:curve_bee}). The metric is Average Precision (AP), based on Object Keypoint Similarity (OKS). It is a metric originally described in the Pascal VOC challenge \cite{everingham2010pascal} and has been widely reported in the human pose literature. The AP we use is an average over multiple OKS thresholds from 0.5 to 0.95, which is the standard for COCO style datasets. There is a parameter $\sigma^2$ in the definition of OKS (eq.~6 in \cite{pereira2020sleap}), which is the standard deviation of human annotator uncertainty for the easiest keypoint. It also implies the tolerance for misaligned keypoints. When the body parts are small, we need a small $\sigma^2$ to indicate a very high precision to align the predicted keypoints with the ground truth keypoints. When the body parts are larger, we can use a bigger value of $\sigma^2$ since the tolerance for annotator uncertainty is larger.

We compare the test APs, which are the AP values evaluated on the test set, for every thousandth iteration for DirectPose, SemiMultiPose-labeled and SemiMultiPose. We initialize all models with the same pre-trained weights. We start running all models with $\mathcal{L}_{supervised}$ (eq.~\ref{eq:sl}) and labeled frames only to make sure that all models are initially trained to approach to good local optima. For SemiMultiPose-labeled and SemiMultiPose, we then introduce the self-supervised loss for labeled frames $\alpha\mathcal{L}_{fl}$ at the 2k'th iteration. Finally, for SemiMultiPose, we incorporate the self-supervised loss for unlabeled frames $\beta\mathcal{L}_{fu}$ at the 5k'th iteration. The late incorporation of unlabeled data ensures that the massive unlabeled data won't mislead the optimization to bad local optima at an early stage. All models are trained till the 15k'th iteration when all optimizations converge. We add the AP values of SLEAP methods to the curves for comparison. Consistent with Fig.~\ref{fig:compare_bee}, SemiMultiPose has the best predictive performance on the test set. Here are a few conclusions we can draw from the results: (1) SemiMultiPose has more significant advantages when the number of labeled frames is small. Train135 has sufficient and saturated information for this honeybee dataset, so the unlabeled frames don't drastically improve the predictions. (2) The blue curves reach plateaus faster than the grey curves. This indicates that SemiMultiPose-labeled improves the convergence rate over DirectPose. It further implicates that the fusion loss for labeled frames can boost the communication between B1 and B3 leading to faster convergence on the test set. 

\paragraph{Mouse pup dataset.} The mouse dataset has one mother mouse and several pups. It is used to study the mother-pup interaction in mouse social behaviors. Each mouse pup has three keypoints that make up its skeleton: head, torso, and rump. There are 798 labeled images and 1,173 unlabeled images for training, as well as 178 test frames. Each labeled image has between 1 and 5 mouse pups. The dataset doesn't have full annotations for all pups in the labeled images. All these images only have at most one pup annotated with a full skeletal structure; other pups have at most one body part annotated. So the labeled training set has a lot of missing values which makes the training much harder. We perform additional preprocessing to help counteract the missing data. First, we select all of the pups in the training set that have full ground truth skeletons. There are 470 such pups in the dataset. Then we crop these pups and overlay 10 of them randomly on a background image, which is just an image from the training set. We create 47 of these images, and use them for training instead of the original data. The intuition is that our model will benefit from seeing more fully labeled pups per image. The pups also have a very similar look with the grit background, which makes detecting them more challenging.

We compare the results on the pup dataset qualitatively (Fig.~\ref{fig:compare_pup}) and quantitatively (Fig.~\ref{fig:curve_pup}, Table~\ref{bee_map}). We use the same metric as previously, and follow a similar training procedure. The main difference is that we introduce the self-supervised loss for labeled frames $\alpha\mathcal{L}_{fl}$ at the 8k'th iteration. Next, for SemiMultiPose, we incorporate the self-supervised loss for unlabeled frames $\beta\mathcal{L}_{fu}$ at the 15k'th iteration. The models are trained till the 40k'th iteration when all optimizations converge. The total number of training iterations is larger than in the honeybee data since this is a noisier dataset and thus more complicated to train. We see that SemiMultiPose is the best method tested for detecting the mouse pups.

\paragraph{Fly dataset.}
The fly dataset was collected for the original SLEAP paper \cite{pereira2020sleap}. Each frame in the videos has two flies presented in grayscale. Each fly has thirteen keypoints on its skeleton: head, thorax, abdomen, two wings, two forelegs, two midlegs, two hindlegs, and two eyes. For our experiments, we divide the 100 labeled images into 40 training frames, 40 unlabeled frames, and 20 test frames.

Next, we compare the results of the fly dataset qualitatively (Fig.~\ref{fig:compare_fly}) and quantitatively (Fig.~\ref{fig:curve_fly}, Table~\ref{bee_map}). The qualitative results look similar from afar, but SemiMultiPose has higher precision of lining up the predicted keypints with the ground truth keypoints when zoomed in. We use the same metric as for the other datasets. We follow the same training schedule as for the honeybee dataset since both of those contain relatively simple, clear images. Once again, we see that SemiMultiPose has the highest predictive power for the fly dataset.

\section{Limitations}
SemiMultiPose exploits unlabeled information in individual frames, but does not take into consideration the spatial and temporal information that is contained in full unlabeled videos. Therefore, DeepGraphPose may outperform SemiMultiPose in the case where there is only one animal and the unlabeled frames come from a sequential video. Additionally, we achieve the greatest improvement 
when the data is sparsely labeled. When there is an abundance of labeled data, our method may not significantly outperform other methods. 

\section{Conclusion}
We propose a novel semi-supervised learning framework for multi-animal pose estimation. The model extends a state-of-the-art end-to-end multi-person pose estimation method, DirectPose, with self-supervsied learning for unlabeled data. We conduct experiments on three animal datasets. SemiMultiPose outperforms two competitive multi-instance approaches in both HPE and APE with better predictive power and convergence, especially in sparsely-labeled regimes.


{\small
\bibliographystyle{named}
\bibliography{ref}
}

\end{document}